%% file: bmvc_final.tex
\documentclass{bmvc2k}

\usepackage{times}
\usepackage{latexsym}

\usepackage{microtype}
\usepackage{xspace}
\usepackage{graphicx}
\usepackage{caption}
\usepackage{multirow}
\usepackage{multicol}
\usepackage{booktabs}
\usepackage{makecell}
\usepackage{tabularx}
\usepackage{amssymb}
\usepackage{xspace}
\usepackage{enumitem}
\usepackage{comment}
\usepackage{fdsymbol}
\usepackage{etoolbox}
\usepackage{amsmath}
\usepackage{bbm}
\usepackage{adjustbox}
\usepackage{pifont}

\usepackage[T1]{fontenc}

\usepackage[utf8]{inputenc}
\usepackage{contour}

\definecolor{Fuchsia}{RGB}{255, 0, 255}

\title{Long Story Short:\\ a Summarize-then-Search Method \\for Long Video Question Answering}

\addauthor{Jiwan Chung}{https://jiwanchung.github.io}{1}
\addauthor{Youngjae Yu}{https://yj-yu.github.io/home}{1}

\addinstitution{
 MIR Lab\\
 Yonsei University\\
 Seoul, Korea
}

\runninghead{J. Chung and Y. Yu}{Long Story Short}

\DeclareMathOperator*{\argmax}{arg\,max} 

\newcommand{\xmark}{\ding{55}}%

 \newcommand\merlottitlefont[1]{{\color{Fuchsia} \textbf{{\smash{{\usefont{T1}{cmtt}{m}{n}#1}}}}}}
\newcommand\merlotfont[1]{\smash{{\usefont{T1}{cmtt}{m}{n}#1}}}
\newcommand{\modelname}{\merlotfont{LSS}\xspace}

\newcommand{\checkmodelname}{\merlotfont{CLIPCheck}\xspace}
\newcommand{\modelnamelong}{\merlottitlefont{L}ong \merlottitlefont{S}tory \merlottitlefont{S}hort\xspace}

\def\eg{\emph{e.g}\bmvaOneDot}

\def\etal{\emph{et al}\bmvaOneDot}

\newcommand{\ie}{\textit{i}.\textit{e}., }

\begin{document}

\maketitle

\begin{abstract}
    \input{sections/00-abstract.tex}
\end{abstract}


\input{sections/01-intro.tex}

\input{sections/xx-method.tex}

\input{sections/xx-experiment.tex}

\input{sections/xx-related-work.tex}

\input{sections/xx-conclusion.tex}

\bibliography{main}

\input{sections/10-appendix.tex}

\end{document}

%% file: sections/00-abstract.tex

Large language models such as GPT-3 have demonstrated an impressive capability to adapt to new tasks without requiring task-specific training data. This capability has been particularly effective in settings such as narrative question answering, where the diversity of tasks is immense, but the available supervision data is small. In this work, we investigate if such language models can extend their zero-shot reasoning abilities to long multimodal narratives in multimedia content such as drama, movies, and animation, where the story plays an essential role. We propose \modelnamelong, a framework for narrative video QA that first summarizes the narrative of the video to a short plot and then searches parts of the video relevant to the question. We also propose to enhance visual matching with \checkmodelname. Our model outperforms state-of-the-art supervised models by a large margin, highlighting the potential of zero-shot QA for long videos.

%% file: sections/01-intro.tex
\section{Introduction}
\label{sec:intro}

\begin{figure*}[t]
\centering
\includegraphics[trim=0.0cm -0.5cm 0cm 0.0cm,clip,width=0.96\textwidth]{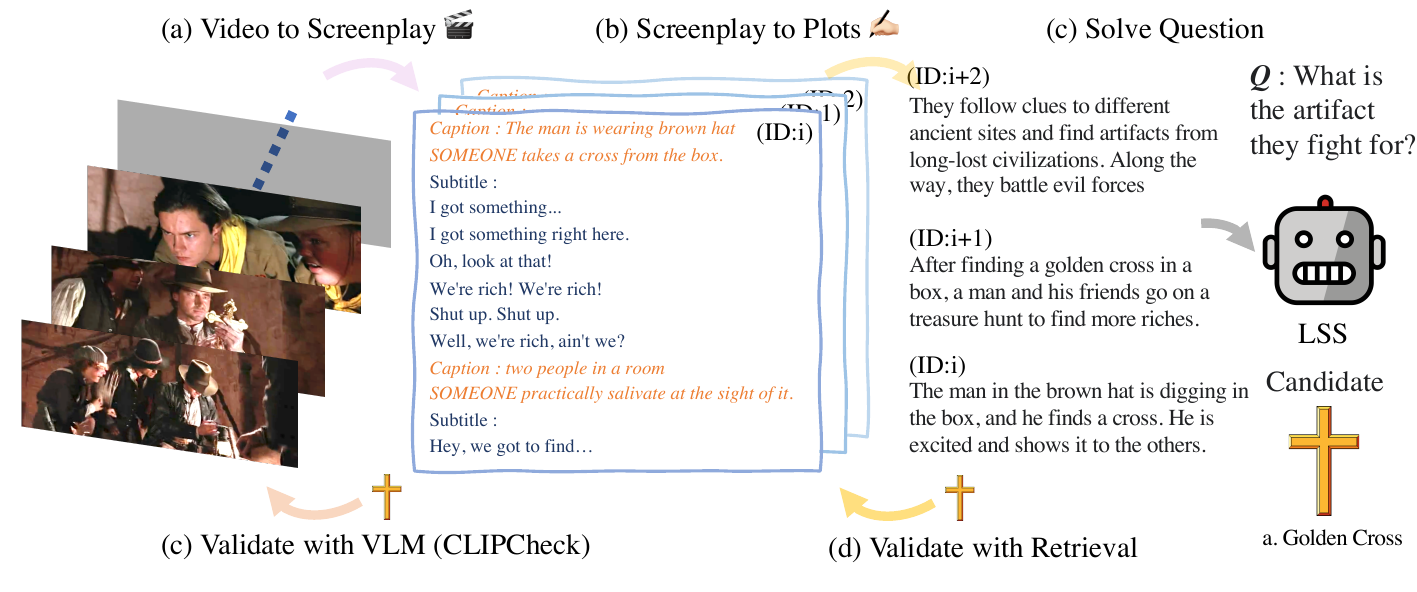}
\caption{\modelnamelong (\modelname) uses Large Language Models (LLMs) (\ie GPT-3) to generate (a) Screenplay and summarized (b) Plots from video. Further details about data processing can be found in Section~\ref{sec:method}. When \modelname answer questions about the video, the model (c) validate given raw video footage with Visual Language Model, CLIP, and (d) search further grounded scripts in a backward manner, which we call \checkmodelname in Section~\ref{sec:visual-checking}.
}
\label{fig:script_idea}
\end{figure*}


Recent video QA models face challenges in handling long video narrative QA tasks~\cite{choi2020dramaqa,tapaswi2016movieqa,kim2017deepstory} (\ie films, dramas, and YouTube web videos) due to the limitation in data and annotations.
This results in an inability to comprehend the long video narratives beyond answering mainly visual questions on short video clip~\cite{xu2017video,lei2018tvqa,lei2019tvqa}. 
The sizes of such long video QAs are insufficient to train the models to fully comprehend the complex narrative structures within a video, yielding sub-optimal performances.
\citep{jasani2019we} demonstrate that the supervised models rely more on language biases in the question than the narrative context:
they can obtain similar performance even without seeing any video context.
This highlights the necessity of multimodal reasoning capability beyond
small task-specific supervision.

To address the challenge caused by low generalization, a zero-shot approach using pretrained Large Language Models (LLMs) can be an efficient alternative for tackling complex QA tasks~\cite{yang2021empirical}, and text context summarization~\cite{zhang2020pegasus,He2022ZCodeAP}. 
Yet, is the narrative QA capability of such LLMs transferable to the video domain?


We propose \modelnamelong (\modelname), illustrated in figure~\ref{fig:script_idea},
that translates video clips into text screenplay format inspired by Socratic Model~\cite{zeng2022socratic}.
Using GPT-3~\cite{brown2020gpt3}, we first summarize the long video into a list of plots and then navigate both the generated summary and the raw video context
to resolve the given question.
Our zero-shot method shows better results than state-of-the-art supervised methods in
MovieQA and DramaQA dataset.
Furthermore, we propose \checkmodelname, a visual-text matching method to enhance visual alignment of the reasoning results provided by GPT-3. 
To summarize, our main contributions are three-fold:
\begin{enumerate}
  \item We present \modelname, a framework that summarizes a long video narrative to a list of plots and retrieves the subplot relevant to the question.
  \item We demonstrate the importance of considering visual alignment strength via CLIP-based matching in visual prompting.
  \item Our zero-shot approach achieves state-of-the-art performance in MovieQA~\cite{tapaswi2016movieqa}
  and DramaQA~\cite{choi2020dramaqa}, outperforming supervised baselines.
\end{enumerate}



%% file: sections/xx-method.tex
\begin{figure*} [t]
\centering
\includegraphics[trim=0.0cm -0.5cm 0cm 0.0cm,clip,width=0.96\textwidth]{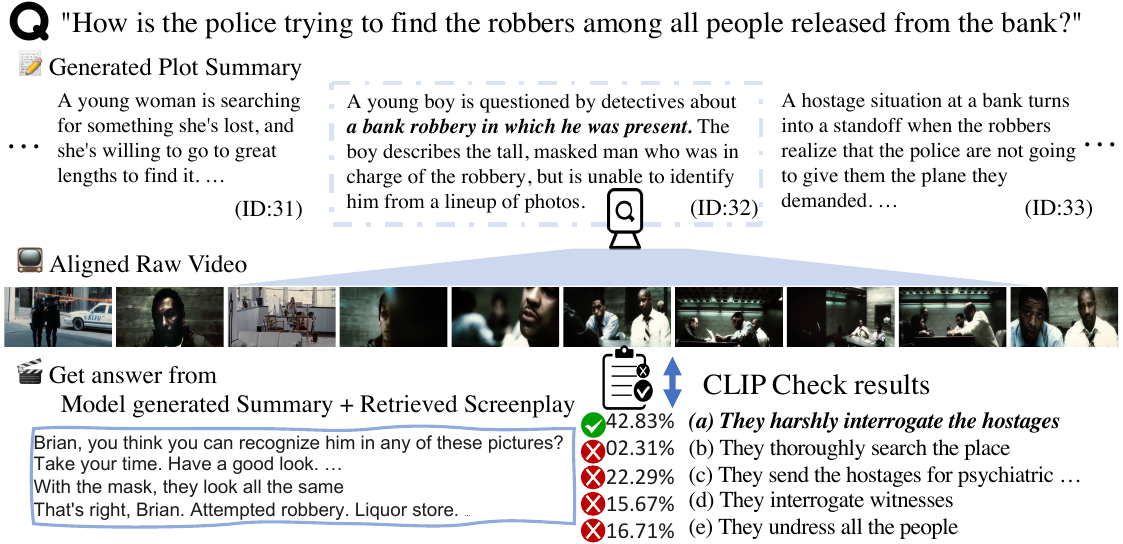}
\caption{The qualitative result showing our proposed \modelnamelong (\modelname) model that generates and retrieves the index of raw video footage. When the model predicts the final answer from (i) the generated Summary and (ii) the retrieved text context, \checkmodelname validates each candidate's answers to revise the final answer for the question. 
}
\label{fig:architecture}
\end{figure*}

\section{Method}
\label{sec:method}

We describe \modelnamelong (\modelname), a prompt-based strategy that divides and conquers narrative video QA by summarizing the plot and retrieving relevant information with the plot. 
Our objective is to predict the correct answer $a_{\hat{y}}$ from long video context $\mathbf{X}$. 
Since language models can only accept tokens of fixed length (\eg 4096 tokens for GPT-3),
they cannot process the rich context (\eg subtitles or scene descriptions) of movies that typically spans two long hours.

Thus, we introduce a summarize-then-search method for video question answering. We first divide the long video context into clip segments and summarize each segment into a subplot using GPT-3 to get the list of subplots $\mathbf{S}_X = \{s_1,s_2 \cdots s_n \}$.
Then, we retrieve video segments $\mathbf{X}_k= \{x_{k_1},\cdots,x_{k_m}\}$ relevant to the question with the subplot list as the input. We use both the full context of the selected segments and the global plot information to derive likelihood over the answer choices $p_\theta (a_1), \cdots, p_\theta (a_5)$.
Finally, we apply \checkmodelname, to strengthen the visual grounding of the selected answer. We provide the prompt templates in the appendix.


\subsection{Plot Generation}
%

We use the ground-truth video partitions to segment the whole video into a set of shorter clips.
Each long video $\mathbf{X} = \{x_1, \cdots, x_n\}$ consists of $n$ clip segments $\mathbf{X_i}$, and each segment contains video $v_i$ and the corresponding text $t_i$ such as subtitle or ASR.
\begin{align*}
     \mathbf{X} = \{(v_1, t_1), (v_2, t_1) \cdots (v_n, t_n)\}
\end{align*}
Given a video $\mathbf{X}$, we first extract visual and text features of the video $\mathbf{X'}$ in natural language forms.
As \citep{yang2021empirical} transcribe image as language prompt for frozen GPT-3, we retrieve 
DVS caption~\cite{rohrbach-arxiv-2016} and image captions with pretrained BLIP~\cite{li2022blip} for video $v_i\in \mathbf{X}$, and concatenate the aligned text $t_i$ as $n$ generated scripts as illustrated in figure~\ref{fig:script_idea}.
We compose a prompt to summarize the context of a video segment into a plot piece of up to three sentences and use GPT-3 to build the plot list aligned with the segment indices
$S_\mathbf{X} = \{s_1, \cdots, s_n\}$. 

\subsection{Narrative Search}
Given the summarized narrative and the question, we wish to retrieve the relatively short clip relevant to the question from the long video.
Language models generate open-ended text which is irregular and often noisy.
To retrieve the exact part of the video,
we drive the model to output indices of the plot rather than the text form. 

We first assign consecutive indices to
the list of summarized plot pieces $S_\mathbf{X} = \{ s_1, \cdots, s_n \} $ aligned with the clip segmentation.
Then, we prompt the language model to output the indices of the plot pieces $k =\{k_1, \cdots, k_m\}$ to lookup for.

The generated indices might still be noisy due to the open-ended nature of language models.
When the model outputs an answer in text form, we use rouge-l~\cite{lin-2004-rouge} score to find plot piece candidates whose similarity with the generated sentence are above the specified threshold $\alpha \ge 0.5$.

Finally, we concatenate the plots $S_\mathbf{X}$, the visual and text representation of the selected segments ${\mathbf{X}'}_k = \{(v_{k_1}, t_{k_1}), \cdots, (v_{k_m}, t_{k_m})\}$, the question $q$, and the answer choices $A = \{a_1, \cdots, a_5\}$ to build the prompt input for question answering.
We process the prompt with the language model with weights $\theta$ and use the index token likelihood as the answer choice score.
\begin{align*}
    p_\theta(a_i) &= p_\theta (i | S_\mathbf{X}, \mathbf{X'}_k, q, A)
\end{align*}


\subsection{Visual Checking}
\label{sec:visual-checking}



For a tighter visual-text matching, we introduce \checkmodelname, a method to conjoin CLIP visual distance~\cite{radford2021learning} and the language model likelihood.
We start from the selected video segments $\mathbf{X}_k = \{x_{k_1}, \cdots, x_{k_m}\}$, answer choices $A = \{a_1, \cdots, a_5\}$, and answer likelihoods $P_\theta = \{p_\theta(a_1), \cdots, p_\theta(a_5)\}$ of GPT-3.

First, we use the CLIP image encoder to encode each frame $x_{k_{i_j}}$ of the selected segments. 
When $l$ is the number of frames within a segment:
\begin{align*}
    \bar{x}_{k_{i_j}} &=  CLIP_V(x_{k_{i_j}}) \\
    \bar{x}_{k_i} &= \{\bar{x}_{k_{i_1}}, \cdots, \bar{x}_{k_{i_l}}\} ,
    \bar{\mathbf{X}}_k = \{\bar{x}_{k_1}, \cdots, \bar{x}_{k_m}\}
\end{align*}

Then, we extract the CLIP text feature of each answer choice $\bar{a}_i = CLIP_L(a_i)$ and compare cosine similarity between the video input and the answers.
We select the best-matched frame for each answer to derive the cosine similarity score.
\begin{align*}
    c(a, \bar{\mathbf{X}}_k) &= \max_{i \le m, j \le l} cossim(\bar{a}, \bar{x}_{k_{i_j}})
\end{align*}

Then we apply the softmax function with temperature $\tau$ on the scores to get normalized visual likelihood over the answer candidates $P_c = \{p_c(a_1), \cdots, p_c(a_5)\}$.
Lastly, we multiply the answer likelihood from the language model $P_\theta$ with the visual likelihood $P_c$ to obtain the final likelihood. We simply select the answer with the maximum value as the model answer.

We choose to consider \checkmodelname only when the language model is not certain of its choice.
Given the likelihoods of the top two answers $p_\theta (a_{h_1}), p_\theta (a_{h_2})$ from the language model, we measure the model certainty with binary entropy $E'$ of the re-normalized probability. 
We only use the combined likelihood when the binary entropy is greater than the given threshold $E' \ge 0.4$.
Otherwise, we do not apply \checkmodelname and just use the language model likelihood.
\begin{align*}
    &h_1 = \argmax_{i \le 5} p_\theta (a_i), 
    h_2 = \argmax_{j \le 5, j \ne h_1} p_\theta (a_j) \\
    &E' = - p_\theta(a_{h_1}) \log p_\theta(a_{h_1}) - p_\theta(a_{h_2}) \log p_\theta(a_{h_2})
\end{align*}

%% file: sections/xx-experiment.tex
\input{graphics/tables/movieqa_small.tex}

\input{graphics/tables/pororoqa.tex}

\input{graphics/tables/dramaqa_table.tex}

\input{graphics/tables/rebuttal_context_ablation}

\section{Experiments}
\label{sec:experiments}

For all experiments, we use GPT-3~\cite{brown2020gpt3} (\texttt{text-davinci-003}) as the backbone language model.
Unless stated otherwise, we use the ground truth clip boundary to segment the videos.
All \modelname variants do not use any training data and thus are zero-shot methods.

\subsection{Evaluating \modelnamelong}

\textbf{MovieQA}~\cite{tapaswi2016movieqa} is a large-scale QA dataset sourced from 408 movies.
There are multiple sources of information in the dataset; subtitles, scripts, DVS, video clips, and plots.
We report four state-of-the-art supervised baselines; A2A~\cite{liu2019a2a}, PAMN~\cite{kim2019pamn}, UniversalQA~\cite{jasani2019we}, and DHTCN~\cite{liu2020dhtcn}.

Table~\ref{tab:movieqa} shows zero-shot \modelname improves over previous supervised approaches.
Also, Ours-search shows strong performance even without the ground-truth segment index label.
\checkmodelname slightly improves the accuracy in the video split.
However, the difference is marginal since MovieQA often requires character-based grounding rather than general visual matching.
Finally, we experiment with the null hypothesis: No Context tests whether GPT-3 solves MovieQA by simply memorizing every fact. 
No Context performs worse than \modelname, rejecting the null hypothesis.

\textbf{PororoQA}~\cite{kim2017deepstory} is a video story QA dataset built from a cartoon series.
The supervised baseline takes the human-generated plot and the ground truth video segment index, while \modelname+Plot+Search takes neither.

Table~\ref{tab:pororoqa} summarizes our result on the PororoQA dataset.
When using both the ground-truth episode and plots, GPT-3 performs almost on par with the supervised baseline.
Substituting a human-generated summary with a model-generated one results in only a marginal performance drop.
Perhaps intriguingly, the search process works better when using model-generated plots.
We attribute this result to the fact that the human annotations are not designed for episode discriminability.

\begin{figure*}[t]
\centering
\includegraphics[trim=0.0cm -0.5cm 0cm 0.0cm,clip,width=1.0\textwidth]{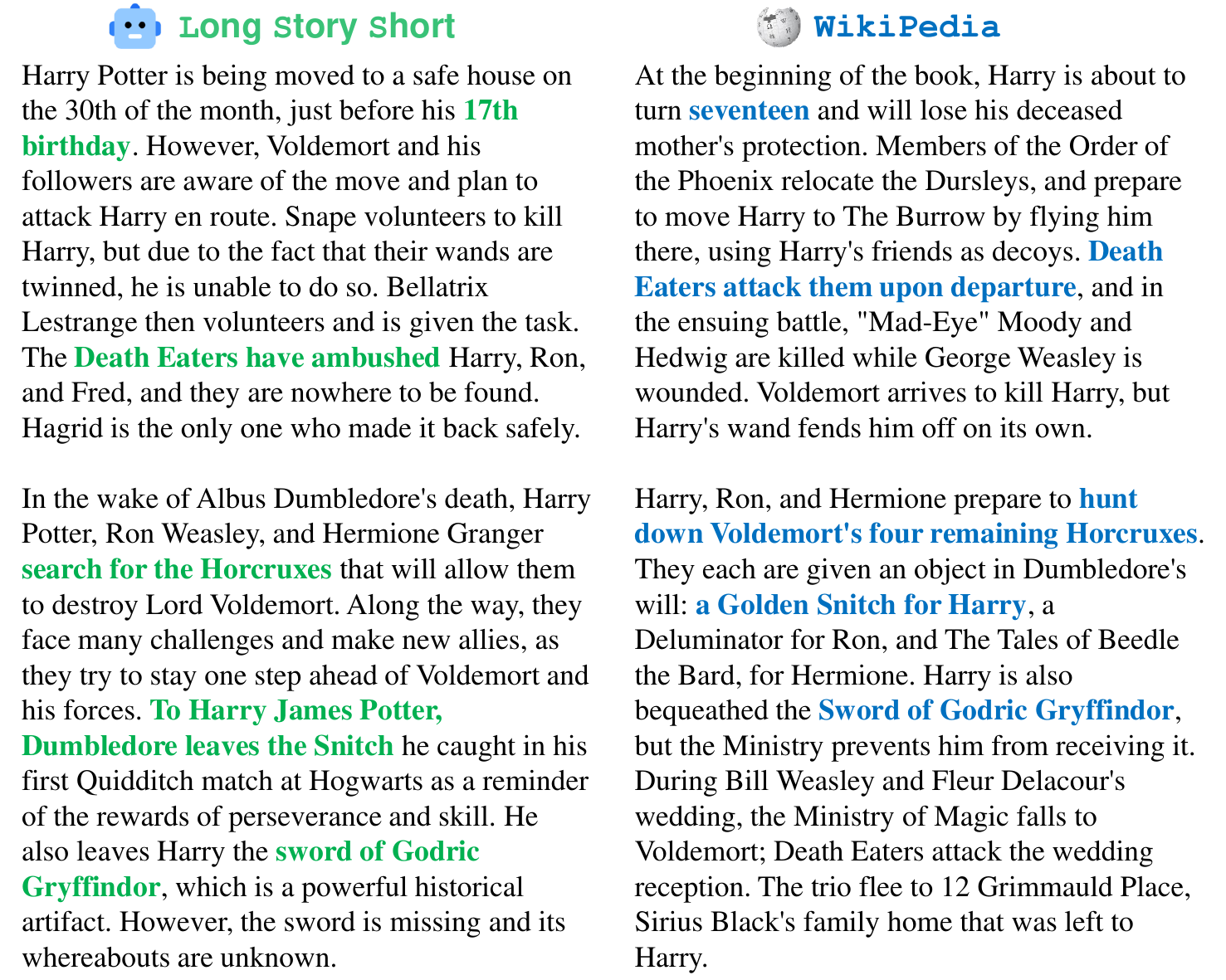}
\caption{Comparison between the plot summary generated by \modelname and the ground-truth summary from Wikipedia. Here, we only show the first two paragraphs of the entire plot because of the space limit.}
\label{fig:plot_sample}
\end{figure*}
\begin{figure*}[t]
\centering
\includegraphics[trim=0.0cm -0.5cm 0cm 0.0cm,clip,width=1.0\textwidth]{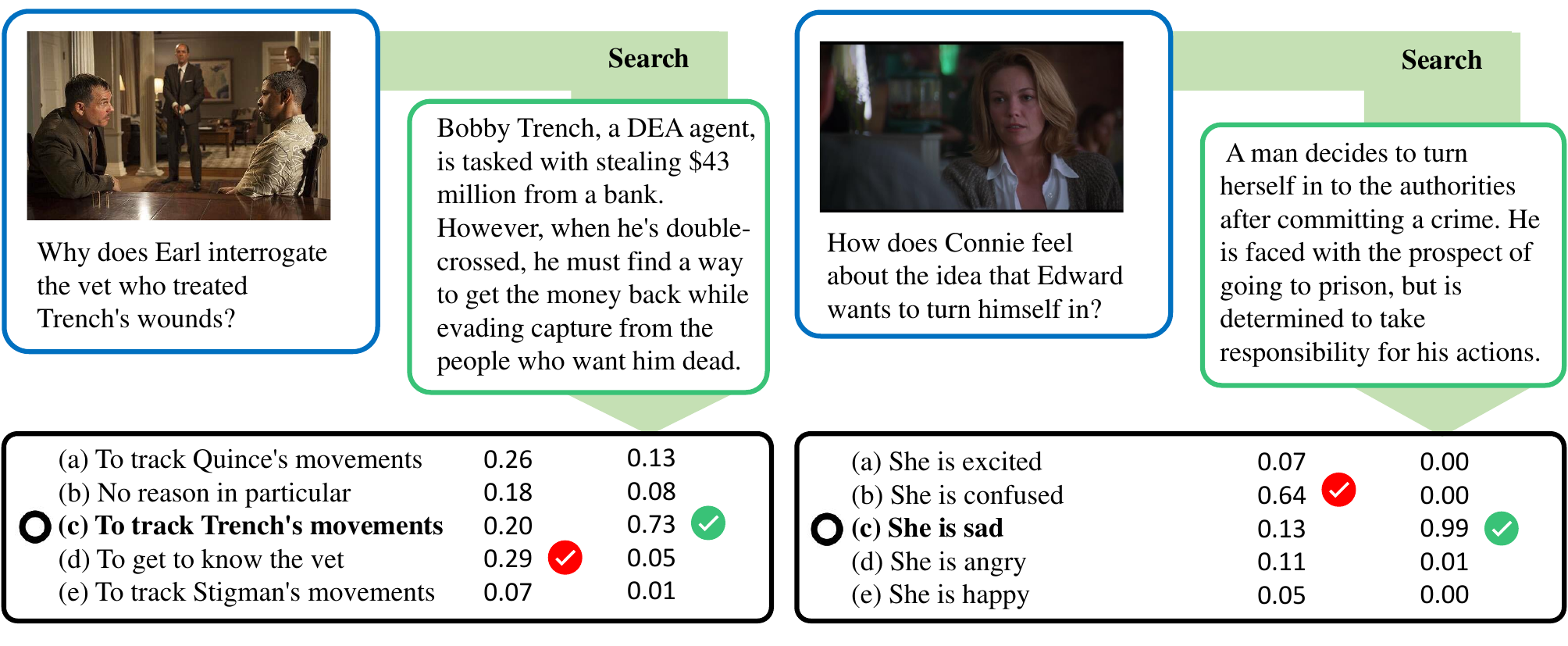}
\caption{QA process samples in \modelname. Conditioning on the searched plot piece has a substantial impact on the language model's answer likelihood distribution.}
\label{fig:plot_sample_qa}
\end{figure*}

\subsection{Evaluating \checkmodelname}

\textbf{DramaQA}~\cite{choi2021dramaqa} is video QA dataset that focuses on story understanding.
The dataset is arranged with four levels of hierarchical difficulty, which follow the human cognitive-developmental stages.
We evaluate \modelname on the two high levels of DramaQA to test plot understanding.
We report two latest baselines in level-wise DramaQA; CharacterAttention and Kim \etal~\cite{kim2021aaai}.

We compare the effect of \checkmodelname and \textit{Caption}, a prompt-based method of incorporating image frame descriptions extracted from BLIP~\cite{li2022blip} as inputs to GPT-3.
Table~\ref{tab:dramaqa} shows that \checkmodelname offers greater improvement than image descriptions. Also, while adding image captions improves \modelname, the gain disappears when used jointly with \checkmodelname.
We suspect that this is because frame captions provide similar information to \checkmodelname while being much noisier.
Note that the automatic \textit{Captions} here are not an integral component of \modelname.
As DramaQA has visually grounded annotations already, adding automatic image \textit{Captions} on top of that would not
necessarily improve the model performance. 
Rather, we use the \textit{Captions} to explicitly compare early vs. late visual alignment methods.

Finally, we check whether \checkmodelname exploits the dataset bias rather than understanding the visual context.
To this end, we devise a variant of \checkmodelname with random visual context (\checkmodelname-Shuffle).
\checkmodelname-Shuffle does not improve over \modelname with no \checkmodelname, denying the bias hypothesis.

\subsection{Ablation Study}

Are both the summarization and search important for narrative understanding?
Here, we evaluate \modelname variants with full context without the narrative search (\modelname-Full) or with the plot summary and random segment as inputs (\modelname-Random).
Table~\ref{tab:rebuttal_context_ablation} shows that both \modelname-Full and \modelname-Random fall behind \modelname-Search, indicating the importance of retrieval.
Note that we could not employ the full context in \modelname-Full due to the token length limitation.
Instead, we use the longest prefix of the full context that GPT3 accepts (4000 tokens minus the length of the instruction).

\subsection{Qualitative Results}

Figure~\ref{fig:plot_sample} shows the automatic plot summary generated as an intermediate context of the long video QA using the language model in the \modelname framework. As shown in the qualitative sample, the generated plots align well with the human-written plots from Wikipedia.
For example, in the first scene of the movie "Harry Potter and the Deathly Hallows", the \modelname summary correctly writes that Harry Potter is currently 17 years old and the main event in which the death eaters attack the protagonist.

Figure~\ref{fig:plot_sample_qa} depicts the connection between the searched plot piece and the answer likelihood.
In the example on the left, the retrieved summary tells that Trench committed a crime and thus is on the run, suggesting that another character interested in him would be chasing him. The language model understands this context to modify the answer likelihood in the correct way.
In the right example, the \modelname plot piece suggests that Edward is confident in his decision. While this context does not offer a direct cue to the question, the language model sees it as information strong enough to alter the answer.

%% file: graphics/tables/movieqa_small.tex
\begin{table}[t!]
\begin{center}
        \begin{tabular}{llcccc}
            \toprule
            & Model & Aligned & V + S
            & V Only & S Only \\
            \midrule                                                                                                                          
            \multirow{4}{*}{\rotatebox[origin=c]{90}{{\footnotesize Supervised}}} &A2A & $\checkmark$    & 41.66     & 40.28    & 41.05 \\
            & PAMN & $\checkmark$    & 43.34     & 42.33   & 42.56                 \\             
            & UniversalQA & $\checkmark$    & 48.87 & 50.67  & 47.62                 \\ 
           & DHTCN & $\checkmark$   & 49.60  & 47.38 & 48.43  \\

            \midrule          
            \multirow{4}{*}{\rotatebox[origin=c]{90}{{\footnotesize zeroshot}}} 
            & No Context & \xmark   & 36.36 & 34.28 & 38.07      \\ 
            & \modelname & $\checkmark$   & 53.44 & 49.83 & 56.42      \\ 
            & \modelname-Search & \xmark  & 51.24  & 49.00 & 53.09    \\ 
            & \modelname-Search+\checkmodelname & \xmark  &  51.49  & 49.55 & 53.09      \\ 
            \bottomrule
        \end{tabular}
    \vspace{0.4cm}
    \caption{Evaluation on MovieQA validation split. The dataset provides GT alignment with 3 minutes of video clip on average: We also report Ours-search which searches the whole movie context without GT alignment. (V) indicates Video and (S) indicates Subtitle. 
    }
    \label{tab:movieqa}
\end{center}
\end{table}

%% file: graphics/tables/pororoqa.tex
\begin{table}[t!] \begin{center}
        \begin{tabular}{llccc}
            \toprule
            \multicolumn{2}{l}{\multirow{2}{*}{Model}} &   \multicolumn{2}{c}{Labels} & \multirow{2}{*}{Acc} \\
            & & Plot & Aligned    \\
            \midrule        
            \multicolumn{2}{l}{Supervised~\cite{kim2017deepstory}}	&  $\checkmark$    & $\checkmark$   & 68.00 \\
            \midrule 
              \multicolumn{2}{l}{GPT3 w\textbackslash o Context} & \xmark & \xmark  & 36.90      \\ 
            \midrule
            \multirow{4}{*}{\modelname} 
              & Base & $\checkmark$ & $\checkmark$  & 66.76      \\ 
              & + Search & $\checkmark$ & \xmark  & 48.98      \\ 
              & + Plot & \xmark & $\checkmark$  & 65.80      \\ 
              & + Plot + Search & \xmark  & \xmark & 53.34   \\             
            \bottomrule
        \end{tabular}
    \vspace{0.4cm}
    \caption{Evaluation on PororoQA validation split. The machine-generated plot (+Plot) performs close to the human annotations (Base).}
    \label{tab:pororoqa}
\end{center}\end{table}

%% file: graphics/tables/dramaqa_table.tex
\begin{table}[t!] \begin{center}
        \begin{tabular}{lccc}
            \toprule
            Model &   Level3 & Level4 \\
                         
            \midrule        
            CharacterAttention	  & 60.82 & 65.62 \\
            Kim \etal~\cite{kim2021aaai}  & 70.00  & 70.00 \\
            \midrule 
            \modelname  & 72.20 &  75.23   \\ 
            +Caption  & 73.54  & 75.68      \\             
            +\checkmodelname  & \textbf{75.78}  & \textbf{79.28}    \\             
            +Caption+\checkmodelname  & 75.34   & 77.93    \\             
            +\checkmodelname-Shuffle & 71.74 & 73.87 \\
            \bottomrule
        \end{tabular}
    \vspace{0.4cm}
    \caption{Evaluation on the levels three and four of DramaQA validation split. \checkmodelname achieves state-of-the-art over the baselines and a prompt-based approach ~\cite{zeng2022socratic} of inputting image descriptions.
    }
    \label{tab:dramaqa}
\end{center}\end{table}

%% file: graphics/tables/rebuttal_context_ablation.tex
\begin{table}[t!]
\begin{center}
        \begin{tabular}{lcc}
            \toprule
             Model & Aligned & V + S \\
            \midrule       
             \modelname & $\checkmark$   & 53.44  \\ 
             \modelname-Search & \xmark  & 51.24  \\ 
             \modelname-Search+\checkmodelname & \xmark  &  51.49   \\ 
            \midrule       
             \modelname-Random & \xmark & 48.92 \\
             \modelname-Full & \xmark  & 48.57 \\
            \bottomrule
        \end{tabular}
    \vspace{0.4cm}
    \caption{Ablation Study on MovieQA validation split.}
    \label{tab:rebuttal_context_ablation}
\end{center}
\end{table}

%% file: sections/xx-related-work.tex
\section{Related Work}
\label{sec:related_work}


\textbf{Movie Summarization}
Movies are typical examples of long videos with clear narrative structures.
Gorinski \etal~\cite{Gorinski2015MovieSS}generate the shorter version of a screenplay as the task of finding an optimal graph chain of a movie scene.
TRIPOD~\cite{papalampidi-etal-2019-movie} is a screenplay dataset containing
turning point annotations. In the same work, an automatic model to identify the turning point from movie narratives is proposed.
Papalampidi \etal~\cite{papalampidi2020ScreenplaySU} later uses the TV series CSI to demonstrate the usefulness of turning points in automatic movie summarization.
Lee \etal~\cite{lee2021TransformerbasedSS} further improves turning point identification with dialogue features and transformer architecture.

\textbf{Long Video QA}
The task of video question answering has been studied extensively in the literature in the form of both Open-Ended QA~\cite{jang2017tgif} and Multi-Choice Problems~\cite{xiao2021next,wu2021star}. Several approaches have been proposed to address this task, starting from RNN-based attention networks~\cite{zeng2017leveraging,xu2017video,zhao2017video,jang2017tgif}, to memory networks~\cite{tapaswi2016movieqa,na2017read,kim2019progressive}, and transformers~\cite{gao2018motion,fan2019heterogeneous}.
Recently, multimodal models pre-trained on large-scale video datasets (VideoQA~\cite{yang2021just},  VIOLET~\cite{fu2021violet}, and MERLOT~\cite{zellers_2021_merlot} and MERLOT-Reserve~\cite{zellers2022merlot})
shows promising performance in video question answering as well.

However, long video QA has received relatively less attention despite its importance. MovieQA~\cite{tapaswi2016movieqa} formulates QAs on the entire movies, which typically span two long hours. DramaQA~\cite{choi2021dramaqa} uses a single TV series as visual context, and tasks a solver to understand video clips 
of length from one to twenty minutes.

%% file: sections/xx-conclusion.tex

\section{Conclusion}

We introduced \modelnamelong, a summarize-then-search method to understand both global narrative and the relevant details for video narrative QA.
Our approach is effective when the context of QA is vast and a high-level interaction with such context is necessary to solve the said QA, which is the case in long video QAs.
Also, we propose to further enhance the visual grounding of the model-generated answer by post-checking visual alignment with \checkmodelname.
Our zero-shot method improves supervised state-of-art approaches in MovieQA and DramaQA benchmarks.
We plan to release the code and the generated plot data to the public.

There are two possible research directions beyond this work: first, providing visual descriptions better aligned with the story with character re-identification and co-reference resolution improve input quality to GPT-3.
Second, one can devise a more dynamic multi-hop search that combines global and local information in a hierarchical manner.


\section{Limitations}

\label{sec:limitations}

Our study has some limitations, including:
\begin{enumerate}[leftmargin=*,topsep=0pt,itemsep=-1ex,partopsep=1ex,parsep=1ex]
\item We experiment with only videos with English subtitles. However, our method can be extended to include multi-lingual contexts given a strong multilingual language model.
\item The computation and memory requirement of our method is substantial due to its heavy reliance on the large language model, GPT-3.
\item We evaluate \modelnamelong with only a single instance of LLM (GPT-3).
\end{enumerate}

\textbf{Potential Risk}.
Summarizing the long video context with GPT-3 carries on ethical risks related to the open-ended nature of the language model.
GPT-3 may (a) hallucinate fake facts about the content, (b) generate toxic utterances, or (c) implicitly embed social biases into the summary and the answer likelihoods.

%% file: sections/10-appendix.tex
\clearpage
\appendix

\section{Experiment Details}

\textbf{Computational Budget}.
 \modelnamelong uses GPT-3 (175B parameters) via OpenAI API as the backbone.
 An average prompt to summarize a video segment processes $\sim 3000$ tokens,
 while a QA prompt usually takes $\sim 4000$ tokens.
 For \checkmodelname, we extract CLIP features and compute the cosine similarity using a single NVIDIA A6000 GPU: it takes 0.5 hours to process video frames for the MovieQA validation split.

\textbf{Hyperparameters}.
All hyperparameters are pre-defined by analyzing a single training sample.
 For narrative search, we use sentence similarity threshold $\alpha \ge 0.5$ to find plot pieces when GPT-3 does not output a single index.
We use the binary entropy threshold $E' \ge 0.4$ in \checkmodelname.
We run each experiment only once, as our method is deterministic and is not susceptible to randomness in initialization.

\textbf{Video Segmentation Scheme}.
There are predefined segment boundary annotations for all datasets we utilize in this paper.
Also, all plot pieces have aligned clip segments in turn since we perform summarization on each clip segmented with the predefined boundaries. 
Also, before applying \modelname we filter out clip segments that 1. are too short, 2. have no aligned image frame, or 3. have no text context to make sure that we can retrieve the clip segments using plot summaries.

\textbf{External Libraries}.
We use OpenAI API to access GPT-3 language model. 
The CLIP features are computed with the Huggingface implementations (\url{https://huggingface.co/docs/transformers/main/en/model_doc/clip}).

\section{Prompt Samples}

We use the following prompts for each stage of \modelnamelong.
We break lines for visibility and instead denote the actual linebreaks with \texttt{\textbackslash n}.
Also, listed items within the prompts are abbreviated using ellipses (\texttt{...}).

\textbf{Screenplay to Plot}.
\begin{verbatim}
I am a highly intelligent storytelling bot. 
If you give me a script, I will give
you the short synopsis in detail.\n\n
[Generated Screenplay]\n\n
Synopsis:
\end{verbatim}

\textbf{Plot Index Lookup}.
\begin{verbatim}
Plot:\n
(1) [Plot1]\n
(2) [Plot2]\n
...\n
(N) [PlotN]\n\n
I am a highly intelligent question answering bot.
If you provide me with a question, I will give you
an index of the plot you should lookup to solve it.\n
Q: [Question]\n
Top 1 Plot Index: (
\end{verbatim}

\textbf{Answering Question}.
\begin{verbatim}
Plot:\n
(1) [Plot1]\n
(2) [Plot2]\n
...\n
(N) [PlotN]\n\n
[Generated Screenplay]\n\n
I am a highly intelligent plot question answering bot.
If you ask me a question and candidates, I will give you
the index of answer.\n
Q: [Question]\n
Candidates:\n
(1): [Answer1]\n
...\n
(5): [Answer5]\n
A: (
\end{verbatim}